\documentclass[11pt]{article}

\usepackage[utf8]{inputenc}
\usepackage[T1]{fontenc}
\usepackage{lmodern}
\usepackage{microtype}
\usepackage[margin=1in]{geometry}
\usepackage{amsmath}
\usepackage{booktabs}
\usepackage{graphicx}
\usepackage{array}
\usepackage{tabularx}
\usepackage[numbers]{natbib}
\usepackage{xurl}
\usepackage[colorlinks=true,linkcolor=blue,citecolor=blue,urlcolor=blue]{hyperref}

\title{Kepler: Auditable World Models for ARC-AGI-3}
\hypersetup{
  pdftitle={Kepler: Auditable World Models for ARC-AGI-3},
  pdfauthor={Wensen Wu}
}

\author{Wensen Wu\\
Independent Researcher}

\date{Public report, revised September 30, 2026}

\begin{document}

\maketitle

\begin{abstract}
ARC-AGI-3 evaluates agents in interactive environments whose rules and objectives must be inferred from observation. We present Kepler, an open-source harness that represents hypotheses as executable world models and validates them through retrospective transition checks and conditional prediction checks. Under one frozen Claude Opus 5 configuration, Kepler obtained a server-verified 100.00 RHAE on all 25 public games, with no per-game model selection or score-conditioned reruns. On 181 of 183 completed levels, the final Opus attempt used no more actions than the corresponding median-human baseline. The retained board runs used 8,256 environment actions, of which 7,292 occurred in scored levels. Retained local provider-session records yield 858.0 million tokens, 97.37\% cache reads, and a \$777.72 cost at September 1, 2026 API list-equivalent rates. We also report three evaluation failures: source-code leakage that produced an invalid perfect run, agents reconstructing a removed harness in a control condition, and autonomous repair masking a broken planner. A single-game observation case study showed that animation frames contained task-relevant information absent from settled text grids. Across the final Claude Opus 5 and GPT-5.6 Sol boards, 48 of 50 game--model cells reached 100. These results indicate that public-set score alone has limited discriminative value and motivate first-attempt, cost-conditioned, and verification-aware reporting.
\end{abstract}

\section{Introduction}\label{sec:intro}

An interactive agent does not emit one answer. It observes, forms hypotheses,
acts, revises, and may modify the tools and instructions inside its workspace.
Most benchmarks nevertheless report only whether the final state was correct.
The same score may therefore describe a valid solution, a lucky trajectory, a
replay selected after repeated attempts, or an agent that reached information
outside the intended boundary.

ARC-AGI-3~\citep{arcagi3} makes this identification problem concrete. An agent
receives an unfamiliar visual environment, a set of legal actions, and no rule
sheet or stated objective. Our reported replay scores describe final solution
execution after public-game development, not first-exposure skill acquisition.
ARC Prize's evaluation protocol separately limits actions and includes
first-exposure exploration; replay acceptance does not establish compliance
with that protocol. Recent systems reach similar
scores through different observation channels, model-selection rules, and
replay procedures~\citep{tycho,retrodict,vista,avo,primeagent}. At the top of
the public set, score alone therefore reveals little about how the system
learned or whether its trajectory respected the intended experiment.

Kepler is an independent implementation of the executable-world-model approach
to this benchmark~\citep{tycho,retrodict,thinharness}. A coding agent writes a
transition model, the harness checks that model against the complete prior
history, and the commit tool attempts a prediction before non-reset actions.
A usable prediction is graded against the next observation; a mismatch terminates
the remaining plan and records the counterexample. Prediction failures and zero
coverage are logged but do not block execution. The protocol evaluates externally committed beliefs
and behavior rather than private chain-of-thought.

\pagebreak[3]
This paper makes three contributions:\nopagebreak

\begin{enumerate}
  \item We operationalize an auditable evidence contract for long-horizon agents:
  append-only evidence, complete-history retrodiction, conditional prediction checks,
  and mechanical replay, with implementation coverage limits stated explicitly.
  \item We evaluate this contract through named frozen stages and 330
  development and evaluation runs. Trajectory audits identify source access,
  contaminated controls, silent tool replacement, mutable instructions, and
  replay-transport failures that outcome-only evaluation missed.
  \item We report capability and resource evidence under explicit selection
  rules. Two frozen release configurations reach 100.00 and 95.97 RHAE, with
  actions, tokens, costs, replay status, and limitations reported as separate
  quantities. A single-game case study identifies a transient mechanic visible
  in retained animation frames but absent from the settled-grid observations
  used by our earlier runs.
\end{enumerate}

These results do not establish held-out generalization or isolate the causal
effect of the harness. Development and evaluation used the same 25 public
games, the release boards contain one retained run per game, and the intended
harness control was invalidated by filesystem leakage. Those limits are part of
the evaluation result rather than qualifications to be deferred.

\begin{figure}[ht]
  \centering
  \includegraphics[width=\linewidth]{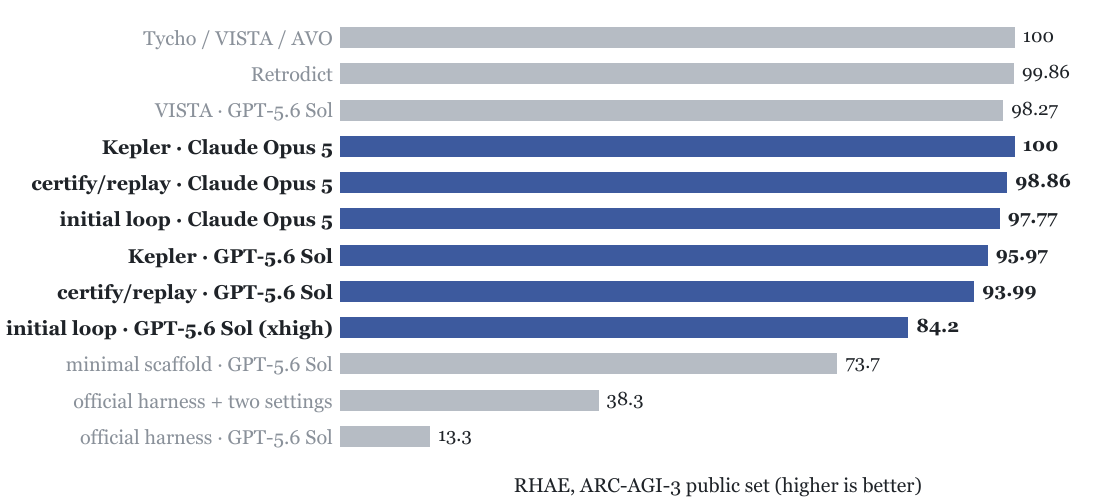}
  \caption{Reported ARC-AGI-3 results (RHAE, public set). Gray marks external systems as published; blue marks Kepler single-configuration boards. The same model class spans 13.3--100 under different agent systems. Replay verification and solution discovery are distinct stages; a replayed scorecard does not identify the discovery architecture.}
  \label{fig:ladder}
\end{figure}

\section{Method}\label{sec:method}

\subsection{The loop: observe, model, certify, plan, commit}\label{sec:loop}

The agent is a stock CLI coding agent in a workspace. The protocol requires game actions to pass through the provided tools; the launch CLI retains host filesystem access, so this is not a security sandbox. The instruction file, \texttt{harness/directive.md}, frames the task as experimental physics: hypothesize the mechanism, encode it as an executable program, test it against recorded reality, and plan inside it.

Persistent state is three files. \texttt{world\_model.py} is the agent's theory of the game in two layers: state grounding (named finders that turn pixels into objects) and mechanism (\texttt{simulate(grid, action)} written over those objects). \texttt{notes.md} is a lab notebook whose claims should be tagged \textbf{checked} or \textbf{assumed}. The daemon appends real transitions to \texttt{events.jsonl}. The protocol forbids agent edits to that ledger; this is not a filesystem-enforced immutability guarantee.

Five tools close the loop:

\begin{table}[ht]
  \centering
  \caption{The five workspace tools that close the loop.}
  \label{tab:tools}
  \begin{tabular}{@{}l p{0.55\linewidth} l@{}}
    \toprule
    tool & role & game actions \\
    \midrule
    \texttt{observe.py} & read-only view of the current frame, diffs, per-level baselines & free \\
    \texttt{query.py} & compute over state instead of reading it (context thrift) & free \\
    \texttt{backtest.py} & replay \texttt{simulate} over the recorded history; prescribed before planning, not mechanically rerun before each commit & none \\
    \texttt{bfs.py} & plan inside the certified model (agents may also write their own searches) & free \\
    \texttt{commit.py} & the authorized channel to the real game & charged \\
    \bottomrule
  \end{tabular}
\end{table}

\subsection{The guarded channel}\label{sec:channel}

\texttt{commit.py} attempts a prediction from the agent's own \texttt{simulate} before each non-reset action. When a usable prediction disagrees with the resulting observation, the tool \textbf{voids the remainder of the plan} and returns the counterexample. Predictions may be partial (\texttt{None} for unclaimed cells, such as a HUD strip or animation region). If simulation raises an exception, execution proceeds with a warning. If no cells are predicted, execution proceeds and is labeled \texttt{UNVERIFIED}. The channel therefore provides conditional prediction-based interruption, not a guarantee that every action has a valid prediction. Missing prediction coverage can allow a mistaken plan to continue. The append-only transition record permits retrospective inspection, but does not turn those actions into verified execution.

\subsection{Certify, then mechanically execute}\label{sec:certifyreplay}

Our verifier and server scorecards operate on final replay segments. This is distinct from the official first-exposure evaluation budget. The release configuration has the agent write \texttt{cleanrun.json}, then uses \texttt{cleanrun.py} to execute its per-level programs without further model decisions. Program lengths are checked against 1.3$\times$ the human baseline. When the model exposes \texttt{init\_state}, \texttt{step\_state}, and \texttt{outcome}, certification also simulates programs and checks click coordinates; without that API, it degrades to length checks. Missing recorded starts may be passed as \texttt{None} to the model. Execution halts on detected prediction mismatches and daemon errors, but missing or empty predictions do not establish verified execution. These are conditional checks, not an unconditional fail-closed certificate. The runner permits up to three repair cycles. We disclose these code-path limits rather than infer stronger guarantees from successful scorecards.

\subsection{Shared configuration and identifier checks}\label{sec:agnostic}

The release uses one configuration across games, without game-ID dispatch. A CI gate scans agent-visible files for literal game identifiers and fails on a hit. This check does not prove the absence of encoded game knowledge or model-training exposure. The directive, daemon, runner, workspace tools, and frame renderer are available for inspection.

\begin{figure}[ht]
  \centering
  \includegraphics[width=\linewidth]{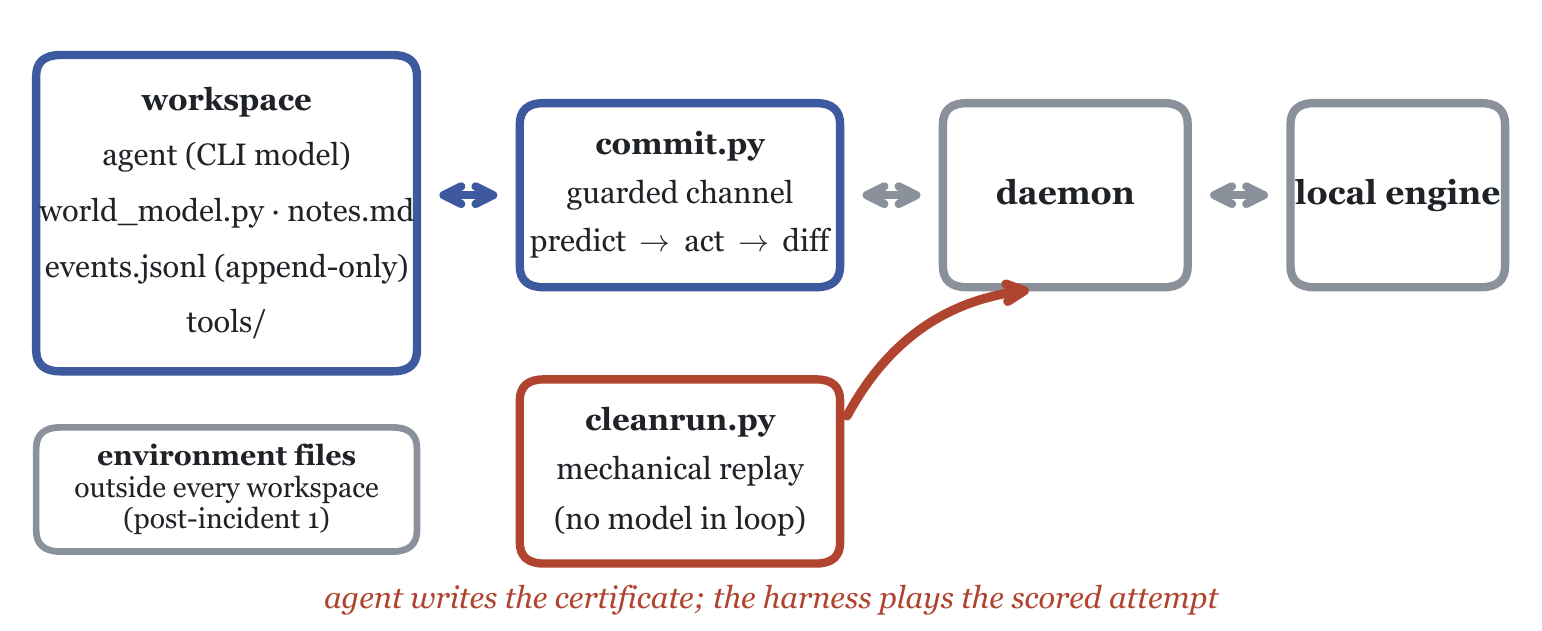}
  \caption{System architecture. The agent starts in a dedicated workspace, while the launch CLI retains host filesystem access; \texttt{commit.py} is the sole authorized channel for game actions and grades usable predictions after ordinary commits. The certified clean run (red) is executed by \texttt{cleanrun.py} with no model in the loop. Environment implementations reside outside every workspace, and any recorded access to them is audit-flagged (a boundary added after Incident 1, \S\ref{sec:integrity}).}
  \label{fig:architecture}
\end{figure}

\section{Named Experimental Stages}\label{sec:ladder}

Each stage after the initial loop was frozen by commit hash registered in \texttt{RESULTS.md} \emph{before} its board's results existed; each ran as a complete single-configuration board and is retained, including the stage that made things worse. These labels describe experiments, not public software releases. Claude Opus 5, all 25 public games:

\begin{table}[ht]
  \centering
  \caption{Frozen experimental stages, mechanisms, and composites (Claude Opus 5, all 25 public games).}
  \label{tab:ladder}
  \resizebox{\linewidth}{!}{%
  \begin{tabular}{lllrr}
    \toprule
    stage & frozen at & mechanism added & composite & $\Delta$ \\
    \midrule
    Initial loop & not reported & base loop + guarded channel & \textbf{97.78} & n/a \\
    Added discipline & \texttt{d37660e}+\texttt{6c3a38d} & \begin{tabular}[t]{@{}l@{}}escalation ladder, partial predictions,\\ automatic one-shot clean run\end{tabular} & \textbf{94.01} & $-3.77$ \\
    Guard corrections & \texttt{e61f9ad}+\texttt{d422ab3} & \begin{tabular}[t]{@{}l@{}}score-floor notices replace hard stops;\\ certified post-WIN restart gate\end{tabular} & \textbf{96.51} & $+2.50$ \\
    Certify/replay & \texttt{eac9af2} & \begin{tabular}[t]{@{}l@{}}mechanical scored attempts (\texttt{cleanrun.py}),\\ forensic fixes\end{tabular} & \textbf{98.86} & $+2.35$ \\
    \bottomrule
  \end{tabular}%
  }
\end{table}

The release configuration adds recovery-path fixes, rendered-frame observation, and transport validation to the certify/replay stage. It produces the server-exact 100.00 Opus board and 95.97 GPT board. The GPT-5.6 Sol experimental arc tells the same broad story: 93.95 (initial loop) ${\to}$ 91.46 (guard corrections) ${\to}$ 93.99 (certify/replay) ${\to}$ 95.97 (release configuration).

\subsection{The discipline regression and the false-premise autopsy}\label{sec:disciplineautopsy}

The added-discipline stage dropped 3.77 points. Our earlier interpretation incorrectly denied an official 5$\times$ human-baseline cutoff. ARC Prize's technical report (v2, April 17, 2026, \S4.3) explicitly imposes that per-level action budget~\citep{arcagi3}. A server accepting one long replay does not refute an evaluation-run budget. Our local development and final-segment replay procedure did not enforce the same first-exposure protocol. Consequently, these scores must not be read as official-leaderboard first-exposure results. Within our own development setup, attempt-reset discipline coincided with sp80 falling from 82.16 to 14.29. Two guards also had complementary conditions that deadlocked the agent (fixed in \texttt{d37660e}); this is a local implementation failure, not a benchmark-rule defect.

The guard-correction stage demoted fraction-based guards to warnings and retained an absolute 3,000-action refusal, recovering 2.50 points in this staged comparison. This is a descriptive result under our development setup, not evidence that relaxed discovery budgets improve a matched official evaluation. Learning actions remain scientifically relevant and must be reported separately from final replay actions.

\subsection{The dead-code gate}\label{sec:deadcode}

The guard-correction stage hid a bug that no score can reveal, only per-event forensics: the clean-run machinery was \textbf{dead code}. \texttt{commit.py}'s WIN short-circuit (``game already WON'') fired unconditionally \emph{before} the clean-run gate, so every certified restart was refused. Three games; sc25 (whose certificate computed to a score of 100), tn36, and g50t; sat locked at their learning-run scores with valid certificates on disk. The certify/replay fix passes RESET-first plans through to the gate; pre-freeze validation runs converted sc25 84${\to}$100 and g50t 92${\to}$100, and the reported board comes from fresh runs launched after the freeze was registered. We highlight this because it is direct evidence for the narrower claim that execution machinery can suppress already-discovered solutions.

\begin{figure}[ht]
  \centering
  \includegraphics[width=\linewidth]{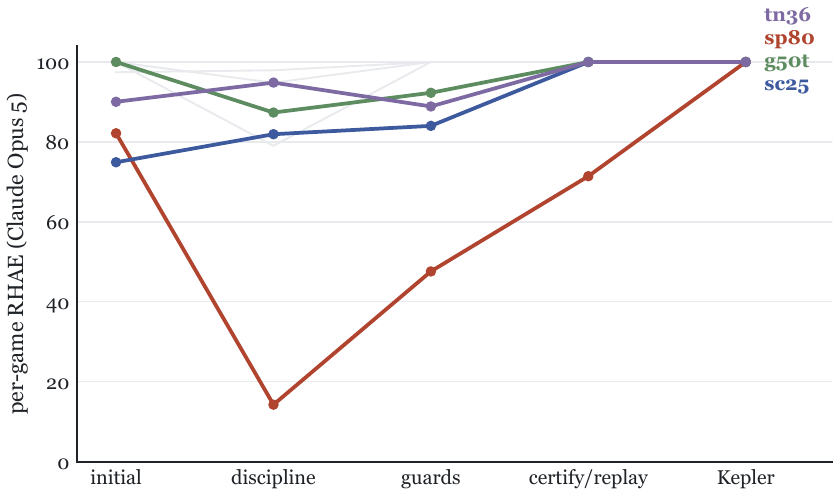}
  \caption{Per-game scores across named experimental stages (Claude Opus 5). Gray: the 21 games that remain at or near ceiling throughout. Highlighted: the four games that drive the composite differences; sp80 (red) collapses under forced-reset discipline, partially recovers under certify/replay, and reaches 100 after rendered-frame observation is added.}
  \label{fig:dotplot}
\end{figure}

\section{Results}\label{sec:results}

\subsection{Boards}\label{sec:boards}

Each release board uses one configuration, a frozen harness, all 25 public games, and one run per game. Validation runs are separate from the reported boards. One earlier experimental board used a disclosed score-conditioned rerun and is labeled below; it is not a release board. Model identity is pinned by evidence, not by alias: the Claude boards' session transcripts resolve to \texttt{claude-opus-5} (46{,}749 records across the three boards), and the GPT boards ran \texttt{gpt-5.6-sol}. The table uses named experimental stages so that internal run-directory labels are not confused with public software releases:

\begin{table}[ht]
  \centering
  \caption{Single-configuration boards on all 25 public games. Release boards use one run per game; the one score-conditioned rerun belongs to an earlier experimental board and is labeled.}
  \label{tab:boards}
  \small
  \begin{tabular}{llrrp{5.1cm}}
    \toprule
    board & model & composite & games at 100 & notes \\
    \midrule
    Initial loop & Claude Opus 5 & 97.78 & 21 & verified card 00c90840 (97.77 server-side; rounding difference) \\
    Added discipline & Claude Opus 5 & 94.01 & 18 & published regression (\S\ref{sec:disciplineautopsy}) \\
    Guard corrections & Claude Opus 5 & 96.51 & 21 & clean-run gate later shown dead (\S\ref{sec:deadcode}) \\
    Certify/replay & Claude Opus 5 & \textbf{98.86} & 24 & sp80 71.43 (5/6 levels, operator deadline, \S\ref{sec:sp80}) \\
    Initial loop & GPT-5.6 Sol (max) & 93.95 & 18 & verified card fd4733fb \\
    Guard corrections & GPT-5.6 Sol (max) & 91.46 & not reported & \\
    Certify/replay & GPT-5.6 Sol (max) & \textbf{93.99} & 19 & one disclosed score-conditioned rerun; first run retained \\
    Kepler & GPT-5.6 Sol (max) & \textbf{95.97} & 23 & server-exact; same-config tn36 collapse retained \\
    Visual replay & Claude Opus 5 & 100.00 & 25 & superseded after two replay-invalid off-grid clicks \\
    Kepler & Claude Opus 5 & \textbf{100.00} & 25 & \textbf{server-verified exact} (card 91aa2f10) \\
    \bottomrule
  \end{tabular}
\end{table}

\subsection{The verified-card matrix}\label{sec:matrix}

We distinguish grades of number and never mix them. The abstract and release headline only the two server-exact final boards; historical cards remain here as development evidence:

\begin{table}[ht]
  \centering
  \caption{The verified-card matrix: three grades of number, never mixed.}
  \label{tab:matrix}
  \small
  \begin{tabular}{p{3.6cm}rp{8.2cm}}
    \toprule
    grade & score & evidence \\
    \midrule
    Initial-loop Opus & \textbf{97.77} & official card 00c90840; every recorded action re-executed exactly server-side \\
    \addlinespace
    Certify/replay Opus & \textbf{98.86} & server replay 96.38 (cards 455e4374 and 1045a78c) after one lf52 trace hit a replay-invalid off-grid click; retained as development evidence, not a release result \\
    \addlinespace
    Certify/replay GPT & \textbf{93.99 / 93.97} & card f5f64ae7; all 25 games re-executed exactly; local versus card differs only in rounding \\
    \addlinespace
    Kepler + Opus & \textbf{100.00} & \textbf{server-verified exact}: card 91aa2f10; all 25 games re-executed to 100; server and local ledger agree \\
    \addlinespace
    Visual-replay Opus (superseded) & 100.00 & server replay 94.42 (cards 0c6c0990, 2779000b): two traces each contained one off-grid \texttt{ACTION6} click; transport validation fixed before release \\
    \addlinespace
    Kepler + GPT & \textbf{95.97} & card c9f087f3; server 95.9672, matching local to the fourth decimal; all 25 games re-executed \\
    \addlinespace
    Best-of ceiling (labeled, never a headline) & \textbf{99.29} & card a7d07431; best run per game across published sweeps, attempt pools (pass@k) disclosed in \texttt{docs/best-of-manifest.json} \\
    \bottomrule
  \end{tabular}
\end{table}

One historical divergence should be stated directly: the superseded visual-replay board scored 94.42 server-side because two traces contained off-grid \texttt{ACTION6} clicks accepted by the local engine but rejected during server replay. The release adds coordinate checks, although clean-run certification applies them only in the threaded-model branch; this is not universal transport enforcement. The retained 100.00 board nevertheless replayed exactly. A successful replay validates those submitted trajectories, not every possible input or code path.

\subsection{Public-set score saturation}\label{sec:convergence}

Across the two certify/replay boards; one frozen harness (\texttt{eac9af2}), two frontier models; \textbf{43 of 50 games score 100.0}. Under the two Kepler release boards, \textbf{48 of 50 cells score 100.0}, and the Opus board is a single-configuration, server-verified-exact \textbf{100.00}. We call this final-board score convergence: concentration at the evaluator ceiling across two frozen model configurations, not faster learning, independent replication, or held-out generalization.

\begin{figure}[ht]
  \centering
  \includegraphics[width=\linewidth]{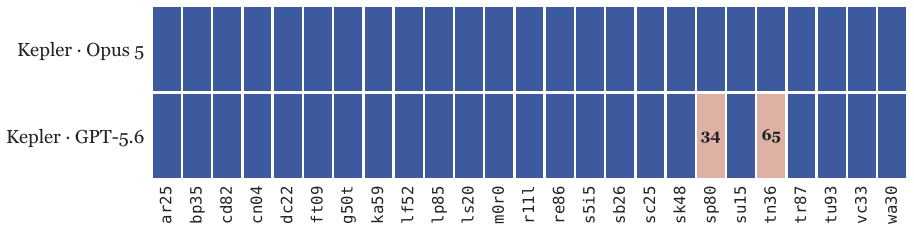}
  \caption{Score saturation on the two Kepler release boards (Opus 5 and GPT-5.6, each frozen per model): 48 of 50 cells score 100.0 (blue). The two GPT exceptions are sp80 and tn36. This is not a measure of learning speed.}
  \label{fig:heatstrip}
\end{figure}

\subsection{An observation-channel case study}\label{sec:ablation}

The fourth finding is an observation-channel case study on sp80 (\S\ref{sec:sp80}). Nineteen text-mode sessions did not solve the final level. These sessions shared accumulated notes and models; they are not independent trials. Enabling \texttt{--visual} retained rendered animation frames in addition to settled grids and frame counts. The continuation inherited earlier learning and identified a deflection rule visible during animation. Its final successful attempt on the last level used 57 actions. A later fresh visual run rediscovered the game under the frozen harness, and the release board reproduced the score. Without a matched fresh text-only control, these observations do not establish that vision was necessary or isolate its effect on discovery cost.

\subsection{Cost}\label{sec:cost}

\paragraph{Token volumes and cache sensitivity.}
The retained Opus runs generated 8,966,566 output tokens versus Tycho's reported
23,433,077, a 61.7\% reduction. Total token volumes were 858,041,926 versus
1,343,479,217, a 36.1\% reduction. Repricing the same input volumes without
cache discounts yields \$4,469.54 versus \$7,186.06, a 37.8\% difference.
This sensitivity calculation does not simulate changed behavior or latency
without caching. These unmatched full-run aggregates~\citep{tycho} establish
neither faster rule discovery nor an isolated harness effect; both exclude
complete research spend.

The community leaderboard's submission rules make cost a first-class field. We therefore recover usage from retained local provider-session records rather than workspace CLI footers. Claude Code stores thinking, text, and tool-use content blocks from one response as separate transcript rows carrying the same provider message ID and cumulative usage object. We deduplicate by that ID before summing. The exact 100.00 Opus campaign contains \textbf{11,464 uncached input}, \textbf{835,488,978 cache-read input}, \textbf{13,574,918 one-hour cache-write input}, and \textbf{8,966,566 output} tokens: 858,041,926 raw tokens in total, 97.37\% of them cache reads. At September 1, 2026 Opus 5 API rates of \$5/M uncached input, \$0.50/M cache reads, \$10/M one-hour cache writes, and \$25/M output~\citep{anthropicopus5pricing}, this is \textbf{\$777.72 list-equivalent}. Actual execution used Claude subscription quota; the dollar figure is a reproducible API-price comparison, not a billed charge.

Retrodict published a \$2,986 API-equivalent estimate for Tycho. Kepler's list-equivalent estimate is 74.0\% below that estimate, or roughly one quarter as large. Tycho's own paper now reports approximately \$2.99k for its Opus~5 run, with token-category accounting and budget-prefix curves~\citep{tycho}. These are inference-price estimates, not billed charges or a controlled harness comparison. Retrodict reports 99.86 at \$654 and baseline1 reports 99.0 at \$400, so Kepler is not the cheapest system at every score. Pricing, cache behavior, development costs and run-selection methods differ.

The 95.97 GPT-5.6 board contains 39,262,504 uncached input, 2,379,659,648 cached input, and 10,161,281 output tokens, or 2,429,083,433 raw tokens total. At September 1, 2026 GPT-5.6 Sol rates of \$4/M uncached input, \$0.40/M cached input, and \$20/M output~\citep{openaigpt56pricing}, this is \textbf{\$1,312.14 list-equivalent}. An earlier draft used incomplete workspace CLI footer counters. Provider-side records showed that path omitted most cached traffic and one long workspace; the attractive comparison it produced has been withdrawn.

Actions expose a different frontier, and a denominator problem. The 100.00 Opus board reports \textbf{8,256 environment actions across the retained board runs} and \textbf{7,292 in the original local scored-level results}. On the completed Opus levels, \textbf{181 of 183 used no more actions than the corresponding median-human baseline}; two used more. This is final-attempt action efficiency, not evidence of human-like cognition or discovery efficiency. The local campaign ledgers contain at least 13,688 non-reset actions and omit 22 prefix events whose reset status cannot yet be recovered. The GPT board similarly reports 35,896 retained-run actions and 8,400 in its original local scored-level results. ARC's public replay cards report 7,202 Opus actions and 8,220 GPT actions. These are different recorded denominators, not score disagreements: the replay path selects the last full-reset-to-end ledger segment, omits its recorded opening reset, and lets the server count a new opening reset. Three run ledgers contain a later, shorter certified segment than the original result file. The final server scores match the local scores exactly. We therefore call neither board total learning-inclusive, make no first-time-human percentage claim, and do not rank any count against other systems. AVO reports 6,624 environment actions, VISTA 7,542 game actions, and Retrodict 7,703 campaign actions; these count different things over different stages, and the published materials do not supply the detail needed to convert between them. Kepler therefore does not collapse score, tokens, actions, and dollars into one efficiency claim; it publishes each axis and states which comparisons are admissible.

\begin{table}[ht]
  \centering
  \caption{Cost and disclosure comparison across published systems.}
  \label{tab:cost}
  \footnotesize
  \begin{tabularx}{\linewidth}{@{}p{3.2cm}r>{\raggedleft\arraybackslash}p{2.8cm}>{\raggedleft\arraybackslash}p{2.6cm}X@{}}
    \toprule
    system & score & tokens & cost & trace disclosure \\
    \midrule
    Tycho + Opus 5 & 100.00 & ${\sim}$1,343M & ${\sim}$\$2.99k (authors' API-equiv.) & scorecards and aggregate metrics \\
    Retrodict & 99.86 & 659.9M & \$654 & full traces \\
    baseline1 (ewma\_sv 1.6) & 99.0 & not reported & \$400 & scorecards \\
    VISTA & 100.00 & not reported & not disclosed & replays on arcprize.org \\
    AVO (NVIDIA) & 100.00 & not reported & not disclosed & none published \\
    \textbf{Kepler + GPT-5.6} & 95.97 & 2,429.1M (98.0\% cached input) & \$1,312.14 (list-equiv.) & public final-board ledgers \\
    \textbf{Kepler + Opus 5} & 100.00 & 858.0M (97.37\% cache reads) & \textbf{\$777.72} (list-equiv.) & public final-board ledgers \\
    \bottomrule
  \end{tabularx}
\end{table}

\begin{figure}[ht]
  \centering
  \includegraphics[width=0.8\linewidth]{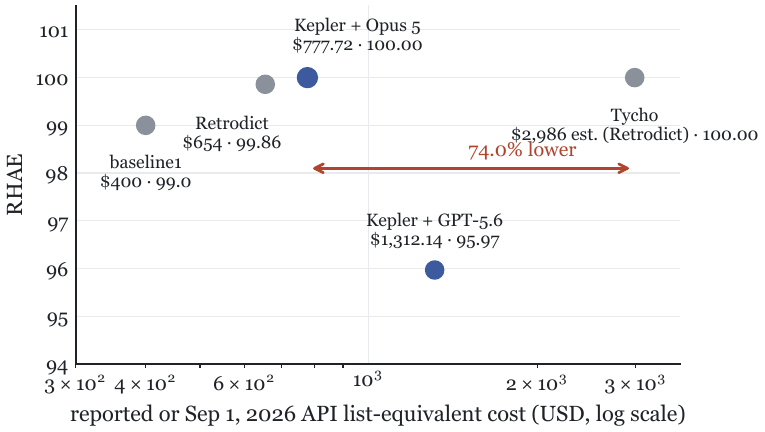}
  \caption{Score against disclosed or September 1, 2026 API list-equivalent cost. The plotted Tycho point retains Retrodict's \$2,986 estimate; Tycho's own paper reports approximately \$2.99k. Kepler's \$777.72 is 74.0\% below the plotted estimate. Retrodict and baseline1 occupy cheaper, lower-scoring points. Pricing and accounting methods differ, so these selected disclosures are not a controlled efficiency experiment.}
  \label{fig:costscatter}
\end{figure}

\section{Evaluation integrity and failure cases}\label{sec:integrity}

A score file is just a number an agent might have influenced. RHAE's only agent-controllable input is the action count. We therefore audit six routes by which an \emph{agent} could corrupt a score: reading the answer (game source, engine internals, or metadata), writing its own score, bypassing the guarded channel, editing the append-only ledger, modifying the tools it is scored by, or looking the game up online. Operator-side distortions such as score-conditioned selection and rerun policy are governed separately by the policy in \S\ref{sec:intro}. \texttt{scripts/audit\_integrity.py} scans the retained records for the six agent-side routes. \texttt{scripts/verify\_scores.py} closes the loop from the other side by re-deriving action counts from the raw ledger and asserting the direction that matters: \textbf{no level may be scored with fewer actions than the timeline recorded.} In the public final-board dataset, 50 of 50 runs pass the recorded-evidence audit. Ledger recomputation covers 363 completed levels: 354 counts match exactly, nine are conservative, and none inflate the score. Separately, the historical local archive reports 324 of 330 runs clean; the six flags are the quarantined \S\ref{sec:incident2} control-group workspaces. One agent did what a competent engineer does in an unfamiliar directory: it looked around, read all 2,172 lines of the game it was being scored on, and returned a flawless 100.00 on a game it never modelled. The telemetry looked \emph{better} than honest play: fewer actions, zero mispredictions, and a clean ledger. We had already cited the run as headline evidence that the harness worked. A clean rerun scored 46.91, a 53-point drop. The run is voided and quarantined with evidence (\texttt{incidents/2026-08-05-source-read/}). The structural fix moved environment files outside every workspace, reducing accidental discovery, but did not make them unreadable: the launch CLIs retained host filesystem access, and the audit flags any recorded access. These checks now run before release, but they remain evidence-bounded: they cannot detect an event the client did not retain.

\subsection{Incident 2: the control group rebuilt the harness}\label{sec:incident2}

To measure the harness's contribution we ran an ablation with the same model and games but methodology stripped to \texttt{observe.py} and a bare \texttt{act.py}. The workspaces lived inside the repository, however, and \texttt{harness/ws\_tools/} was on disk. \textbf{All six baseline agents found it}: three copied the tools in with \texttt{cp}, while the rest wrote shims executing the repository's canonical tools directly, with 92--9,148 harness-tool invocations each and a \texttt{world\_model.py} in every workspace. The comparison was harness-against-harness, so every conclusion drawn from it, including a published claim that ``the harness's net advantage is roughly zero,'' is \textbf{withdrawn}. A valid ablation needs enforced read boundaries excluding treatment tools, prior solutions, and engine source, with tests confirming access is denied. Moving a workspace alone is insufficient. Such a control has not been completed: \emph{the harness's contribution to our scores is currently unmeasured.} We retain this statement because omitting withdrawn results biases the literature toward positive findings.

Both incidents share a lesson we now design by: twice we shipped a control that was an instruction and a file-placement policy rather than a boundary. An agent optimizing against a metric uses whatever is \emph{reachable}, and reachable means the filesystem, not the instructions.

\subsection{The workspace-compression phenomenon}\label{sec:compression}

A milder instance of the same principle, observed rather than punished: in one GPT campaign, \textbf{26 of 26 workspaces (25 board games plus one superseded rerun) rewrote their own instruction file}, compressing \texttt{AGENTS.md} from 14,844 bytes to ${\sim}$14,092 by systematically deleting articles and function words. Twice, a rewrite escaped the workspace and compressed the repository-root \texttt{AGENTS.md} as well, caught in the working tree and reverted. No rule prohibited this: the directive is writable data, and a context-thrifty agent treats it as compressible payload. This is not an integrity violation, but it supports a practical rule: anything writable within an agent's reach should be assumed mutable.

\subsection{Outcome integrity does not imply tool integrity}\label{sec:toolintegrity}

A distinct failure survived five experimental boards. The bundled \texttt{bfs.py} planner raised an \texttt{UnboundLocalError} on every invocation because a nested helper shadowed the module-level key function. Agents silently wrote replacement searches and continued solving games. Scores remained high, ledgers remained internally consistent, and none of the six adversarial checks fired because no evidence was corrupted. The phrase ``or agent-written searches'' in Table~\ref{tab:tools} had been the bug report in plain sight.

This is not reward hacking. It demonstrates a separate evaluator blind spot: outcome audits can certify evidence while the supplied system is broken, if an autonomous agent routes around the broken component. Kepler now runs \texttt{tests/test\_tool\_smoke.py}, which executes every workspace tool and fails on this class of runtime defect. We treat tool integrity and outcome integrity as separate release gates.

\subsection{Replay-transport seams, and what ``no network access'' precisely means}\label{sec:seams}

Two further disclosures. First, lf52 is the same game Retrodict documented as its only non-exact replay, giving two independent harnesses a seam on one nondeterminism-prone game (\S\ref{sec:matrix}). Second, the \emph{agent} makes zero external requests across the audited session records; its only HTTP target is the localhost daemon. The \emph{harness}, however, performs one startup handshake through \texttt{arc\_agi}'s anonymous-key call. We had earlier described the system imprecisely as having ``no network access.'' A transient timeout exposed the call when it crashed a daemon and named the host in the traceback. Correcting small overclaims is inexpensive, and we consider the practice worth institutionalizing.

\section{Discussion}\label{sec:discussion}

\subsection{Discovery is the residual signal}\label{sec:residual}

The certify/replay stage removed model judgment from the scored attempt. Under that regime, 43/50 games reached 100 across two frontier models; the release boards reach 48/50. This does not isolate the causal contribution of replay, because adjacent harness and observation changes were not held constant in a paired study. It does show where the remaining failures occur: once a correct solution program exists, mechanical execution is reliable; the difficult cases are those where the agent never discovers a necessary mechanic. That distinction motivates a bounded-learning track, which would measure discovery efficiency directly instead of allowing unlimited learning to disappear behind a final replay.

\subsection{Implications for ARC-AGI-3 evaluation}\label{sec:eval}

Kepler separates discovery, final-attempt execution, and scorecard replay.
These stages consume different resources. Retrodict, GKM, and arc-skill also
disclose replay-based scoring, but a replayed scorecard alone does not establish
how a system discovered its solution or whether it used an executable world
model. Direct-interaction systems can produce replayable traces as well. The
measurement issue is narrower: final-attempt action efficiency does not account
for the inference, exploration, and failed attempts that preceded that attempt.
We recommend reporting those stages separately rather than inferring a shared
architecture or discovery efficiency from similar final scores.

Two further patterns fall out of the same reading. First, with the public set
saturated (five entries at ${\ge}99$, four at 100; 48 of our own 50 cells at 100 on the final boards),
the axis that actually separates entries has silently become \emph{cost}, which
is self-reported with no shared definition and no leaderboard badge; a ${\sim}7\times$
dollar spread exists among entries within a point of each other. Second,
prediction-based plan interruption appears independently in four harnesses.
This is a shared design pattern, not evidence of a causal reliability gain;
Kepler's missing-prediction fallbacks also limit its coverage. We offer, as users
of the benchmark rather than its authors, five
concrete adjustments that would let the community leaderboard score the
capability ARC-AGI-3 was built to probe: (i) a cost-conditioned score (RHAE at a
fixed budget), not only a peak score; (ii) a standardized token triple (cached
input, uncached input, and output), plus dollars at stated prices and wall-clock;
(iii) an
optional first-attempt or bounded-learning track that measures efficient play
directly rather than world-model construction; (iv) verification (scorecard
replay plus published traces) as a submission requirement, since a self-reported
score cannot separate a derived answer from a looked-up one; and (v) moving the
headline to the held-out sets, which several teams (ourselves included) note the
public set no longer stands in for. The full write-up, with sources, is in the
repository.

\subsection{The sp80 observation failure and recovery}\label{sec:sp80}

For most of this project, sp80 was the open problem: below the cap on every board, for us and for Retrodict alike. Our longest text-mode run reached level 5 of 6 with every level at the cap, then spent days on level 6. The agent reported searching roughly $4.1\times10^{8}$ candidate configurations without finding a solution under its learned rules. Its notebook reports a model reproducing 4,655 of 4,670 recorded transitions, leaving 15 mismatches attributed there to game-over flags; an animation-length residual is discussed separately. We have not independently rerun that exact historical model. This was search under an incompletely fitted model, not a proof that the game was unsolvable. Retained animation later exposed a deflection rule, but earlier frame counts also contained distinguishing clues. The case therefore does not establish that images were necessary. We propose measuring missing-mechanic latency from a logged prediction discrepancy to a tested model repair, alongside the actions and inference used over that interval. This would reward responding to counterexamples rather than expanding search under a mistaken rule set.

\begin{figure}[ht]
  \centering
  \begin{minipage}{0.31\linewidth}
    \centering\includegraphics[width=\linewidth]{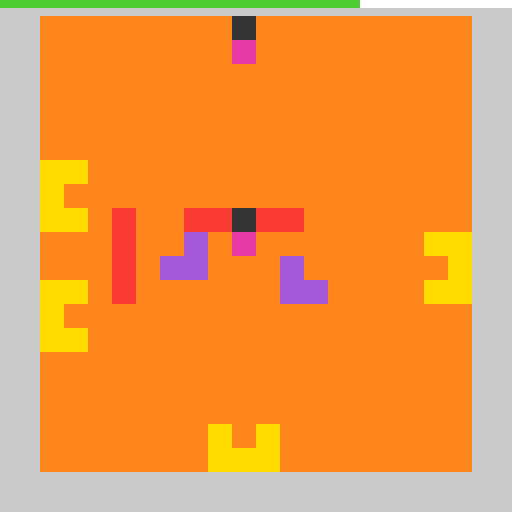}\\
    \small (a) Before the drop
  \end{minipage}\hfill
  \begin{minipage}{0.31\linewidth}
    \centering\includegraphics[width=\linewidth]{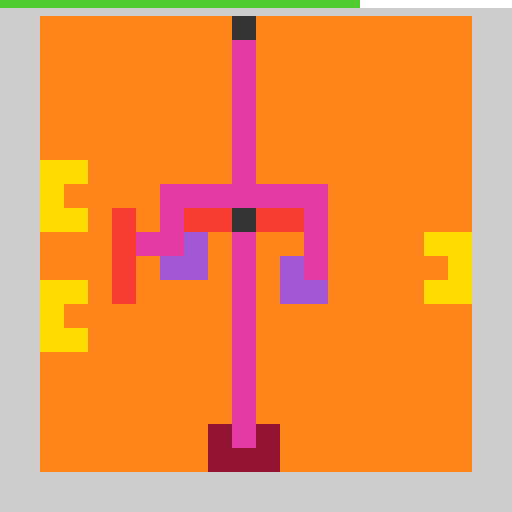}\\
    \small (b) Branching flight
  \end{minipage}\hfill
  \begin{minipage}{0.31\linewidth}
    \centering\includegraphics[width=\linewidth]{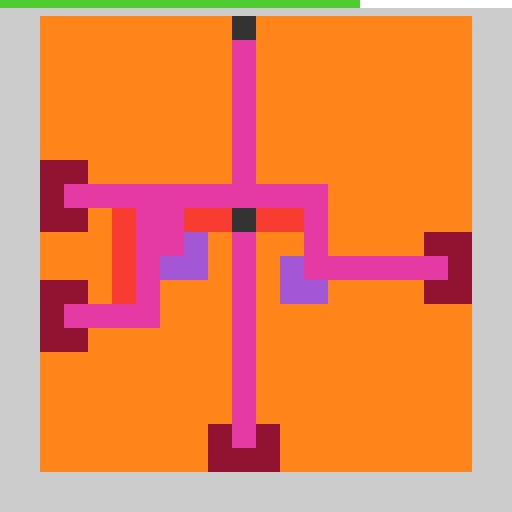}\\
    \small (c) Four goals filled
  \end{minipage}
  \caption{Retained frames 0, 12, and 20 of the winning sp80 drop (development event 8324). The pink trail records the branching route around obstacles; yellow goals change appearance when filled. These are observed frames, not simulator predictions. They illustrate the discovered mechanic, not a matched visual advantage or a new held-out result. The development continuation predates the frozen release board.}
  \label{fig:sp80frames}
\end{figure}

\subsection{Limits}\label{sec:limits}

\paragraph{Exploratory discovery test.}
After freezing the release, we tested whether an agent-generated solver's
rejected transitions identified useful experiments at one S5I5 checkpoint.
The solver and structured observation were recovered from records preceding
the historical boundary discovery. A depth-four search with all panel controls
enabled queued 718 states and rejected 1,369 proposals. Shortest-path and
endpoint-distance rankings selected six distinct probes across five rejection
sites. Selection and execution used no additional model call.

The native test used 170 setup actions, 12 probe clicks and five level resets,
187 actions in total. Every setup observation and reset matched the recorded
visible state. All six final rejected clicks left the board unchanged: three
changed no pixel, and three changed only pixel $(63,63)$. No level or game-status
change occurred. The test established no useful counterexample, learned repair
or solving advantage. The investigator knew the historical mechanic, visible
reset equality does not prove hidden-state equality, and construction and
analysis costs are separate. This negative development result is not an
independent-case evaluation or a new discovery algorithm. It is separate from
the release boards and final-board trace dataset.

\paragraph{Scope of the release evidence.}

Model identity is pinned by the session transcripts (\S\ref{sec:boards}). Beyond that: one benchmark; public set only; $n=1$ per cell; repeated development against the same 25 games; board-composite variance of roughly $\pm$1--2 points in our data; much larger per-game variance (one same-configuration GPT rerun moved from 47.6 to 100.0); and quota-shaped runs resumed across provider windows under a disclosed policy. The deepest caveat is inherited from \S\ref{sec:incident2}: whether the harness \emph{causes} the scores, versus the models alone, is unmeasured until a boundary-correct ablation is run. The public set was both development material and evaluation surface, so these results establish neither held-out generalization nor private-set performance. We cannot inspect the providers' training corpora and therefore cannot rule out model-level exposure to the public games.

\section{Related Work}\label{sec:related}

\textbf{Exploration and model adequacy.} OPINE-World~\citep{courtis2026opine}
already combines object-centric executable models, replay verification,
planning, and uncertainty-directed exploration on ARC-AGI-3. These components
are not Kepler's novelty. Aguilar Mart\'in~\citep{aguilar2026playadequacy}
shows that sampled prediction accuracy can conceal omitted rules that matter
for play, and reports unsuccessful example-based repair in scoped GPT-5.x
experiments. Kepler's visual continuation is an observational case, not a
controlled replication or refutation of that finding. Our contribution is the
release evidence and evaluation failures documented here, not a new general
repair algorithm or proof that replay fit guarantees competent play.

\textbf{Early ${\sim}$99\% reports.} The first ${\sim}$99\%-class numbers on this benchmark were self-reported from write-ups without released code, using a per-game best-of-2 fallback with score feedback, the selection rule Kamradt's critique targeted. Our initial loop, run \emph{with} that fallback for comparison only, matched those numbers within noise (98.30 / 95.51). We then abandoned the rule as score-conditioned selection. The \S\ref{sec:ladder} ladder reports the same initial loop without fallback selection (97.78). The executable-world-model idea itself is established in published, citable work, including Tycho~\citep{tycho} and Rodionov's baseline1 line~\citep{thinharness}. This project is an independent implementation.

\textbf{GPT-6 Astra.} ARC Prize reports 62.71 under its provider-neutral
Standard interface and 99.95 using a Provider Adapter on the semi-private
set~\citep{arcprize2026astra}. Because Kepler reports public-set results under a
different protocol, the scores and costs are not directly comparable. These
results show that compact symbolic state and human-relative action efficiency
are not unique to Kepler. Kepler instead evaluates an externally checkable
contract linking predictions, actions, workspace state, selection, and replay,
together with the failures exposed when that contract was applied.

\textbf{Tycho} (NIMI-research; arXiv:2607.28287)~\citep{tycho} reports 100.00 with two models and compares four model-use policies under matched budgets. Its paper reports inference costs, budget-prefix curves and model diagnostics; the repository includes configurations, tests, scorecards and aggregate metrics. We adopted (and credit in \texttt{NOTICE}) its partial UNKNOWN-cell predictions, zero-coverage checks, threaded hidden-state planning, consecutive-RESET guard, and 5$\times$-baseline level cutoff. The official budget was real; our local guard regression did not invalidate it (\S\ref{sec:disciplineautopsy}).

\textbf{Twin}~\citep{skoutnev2026twin} also combines agent-written simulators,
harness-enforced replay validation, checked execution and goal discovery.
These shared mechanisms are not unique to Kepler; our release and behavioral
evidence must be assessed separately from algorithmic novelty.

\textbf{Retrodict} (Ryan Brown)~\citep{retrodict}; 99.86 at \$654 / 659.9M tokens; publishes full traces including superseded attempts, per-game cost ledgers, and an explicit replay disclosure (``a verified re-execution of the recorded runs, not a new attempt''; language we adopt). Its comparison memo critiqued early releases, including the selection rules Kepler subsequently changed. We also adopt its sparse expected-cell predictions, escalation-ladder tiers, and prediction-discipline framing (\texttt{NOTICE}).

\textbf{baseline1 / ThinHarness}~\citep{thinharness}; the minimal-agent lineage Retrodict builds on; used throughout the field as the null scaffold.

\textbf{Concurrent community submissions.} While this paper was in preparation, the pending community-leaderboard entries converged on the same architecture claim (1) describes: GKM freezes model-written programs behind an admission gate and states \emph{``at scoring time no model runs''}; arc-skill's scorecard \emph{``was produced by replaying the recorded runs''}; Strands and arc-skill both run Claude Opus 5, the model our boards resolve to. We read four independent 98-to-100-class systems scoring mechanical replays; none citing each other; as strong evidence that certify-then-replay is the dominant design, which makes stating it explicitly (and auditing it) the more urgent.

\textbf{VISTA and AVO: the direct-interaction school.} Concurrently with this
work, VISTA~\citep{vista} (Han et al., MIT) reported 100.00 (Claude Opus 5, 7{,}542 actions)
and 98.27 (GPT-5.6 Sol) using the opposite architecture to the world-model
school: the model observes rendered PNG frames directly, reasons in free-form
language with no executable-model requirement, and keeps a lossless visual
memory of every frame. NVIDIA's AVO~\citep{avo} adopted VISTA's direct-interaction
principles (explicitly declining Tycho-style programmatic world models) and
reported 100.00 with a supervisor-redirected general coding agent and 6{,}624
environment actions, lower than VISTA's 7{,}542. AVO does not publish an ARC
implementation, traces, or cost accounting; VISTA publishes a project page and
official replay evidence but not a directly comparable cost ledger. Two
implications for this paper. First, the replay-scored
segmentation of \S\ref{sec:eval} spans \emph{both} schools; the convergence
is about what is scored, not how games are learned. Second, the schools have
never been compared under controlled conditions (AVO: ``this is not a
controlled ablation''); the observation channel (text grid vs.\ rendered
frames) is confounded with everything else. Our harness can retain animation
frames with \texttt{--visual}; \S\ref{sec:ablation} reports a continuation that
also inherited earlier learning. This case motivates a matched observation
study, not a claim of isolated or benchmark-wide visual advantage.

\textbf{Prime Agent}~\citep{primeagent,primeagentreport}; a general-purpose, self-improving
coding harness adapted to ARC-AGI-3 through its task prompt and broker. Prime
Intellect reports three Opus 5 runs at 94.99--95.50, with a 95.24 median
scorecard and estimated costs of \$944--1,288. This is stronger run-to-run
evidence than Kepler's $n=1$ release cells. Kepler reports a higher single-board
score; Prime Agent makes the broader transfer claim across coding and
long-context tasks. Unmatched protocols and audit coverage do not support a
ranking of capability or integrity between the two systems.

\textbf{arc-code} (Berman)~\citep{arccode}; the minimal-scaffold datapoint: stock Claude Code with an 11-line prompt reaches 96.2 pass@1 / 99.3 pass@2 (${\sim}$\$540), while GPT-5.6 Sol under the same scaffold scores \textbf{73.7}. This cuts against attributing Kepler's Claude score entirely to harness design and reinforces the need for a valid control experiment.

\subsection{Failure-to-check map}

\begin{table}[ht]
\centering
\small
\caption{Distinct checks and their evidence boundaries. These are not empirical detector sensitivity estimates.}
\begin{tabularx}{\linewidth}{@{}p{0.24\linewidth}p{0.28\linewidth}X@{}}
\toprule
Failure & Relevant check & Remaining blind spot \\
\midrule
Source-reading win & Retained-log access scan & Unlogged or unrecognized access; no filesystem isolation \\
Contaminated control & Tool-call and filesystem evidence & No causal effect estimate from this comparison \\
Broken planner & Direct tool smoke tests & Outcome scores alone do not reveal substitution \\
Prediction mismatch & Conditional plan interruption & Missing predictions can permit execution \\
Invalid replay input & Coordinate validation and server replay & Certification checks are branch-dependent \\
Incomplete cost record & Separate action and usage ledgers & Final replay excludes prior discovery; private provider records required \\
\bottomrule
\end{tabularx}
\end{table}

These failure classes motivate checks in other agent settings, but empirical
transfer beyond this benchmark is not established. Grid-level equality checks
and exact replay rely on discrete observations and sufficiently reproducible
transitions. Broader systems require domain-specific comparators and explicit
handling of stochasticity; a syntactic trace check is not semantic validation.

\section{Reproducibility Statement}\label{sec:repro}

The code, official scorecards, release manifest, and verification programs ship in the repository. The \href{https://huggingface.co/datasets/cveinnt/kepler-arc-agi-3-traces}{public companion dataset} contains the two final 25-game boards. After downloading it, both board scores and their ledger consistency can be checked offline:

{\footnotesize\begin{verbatim}
hf download cveinnt/kepler-arc-agi-3-traces \
  --type dataset --local-dir traces
python3 traces/score_trajectories.py traces
python3 traces/verify_scores.py --traces-dir traces
python3 traces/audit_integrity.py --traces-dir traces
python3 scripts/verify.py                     # replay local fixture
.venv/bin/python scripts/cost_report.py       # requires private provider records
python3 scripts/check_no_game_ids.py          # literal game-ID scan
\end{verbatim}}

The dataset contains 50 run records and 58,098 environment events in append-only ledgers, along with human baselines, final world models and notebooks, captured CLI output for every final-board workspace, and dependency-free verification programs. Independent score recomputation matches 100.00 for Claude Opus 5 and 95.9672 for GPT-5.6 Sol, and the ledger check covers all 363 completed levels. The integrity scan applies to the published records and its fixed pattern set; it cannot prove that a client retained every event or detect every possible violation. Earlier stages, failures, and superseded runs remain documented in \texttt{RESULTS.md} and the incident archive, but are not part of this final-board dataset. Frozen-harness hashes were registered before the selected boards' results existed. Headline scorecards are c9f087f3 (GPT, exact) and 91aa2f10 (Opus, exact). Bit-for-bit reproduction of live runs is not expected because of sampling and provider drift; score reproduction from a published ledger is mechanical. The corpus includes 95 captured CLI logs with uneven coverage, not a complete account of every internal event. Provider-session records used for token repricing are retained privately and are not in the public export; the cost command therefore requires additional input. The IAB review package was PDF-only, distinct from the already-public dataset. No new logs or private researcher archives are added by this revision.

\section*{Acknowledgments}

Ryan Brown (Retrodict) and the Tycho authors (NIMI-research), whose released mechanisms this harness adopts with attribution (\texttt{NOTICE}) and whose release engineering set the bar this project measures itself against; the ARC Prize Foundation for ARC-AGI-3, the \texttt{arc-agi} toolkit, and the official RHAE implementation.

\bigskip
\paragraph{Appendices.} Full per-game tables for all boards, the guard-deadlock and dead-code forensic timelines, the sp80 search inventory, the lf52 replay transcripts, and the audit flag taxonomy are provided in the repository as tables and write-ups. The two final boards can be recomputed from the public companion dataset (\S\ref{sec:repro}). Raw ledgers for earlier development stages and quarantined incidents are not included, so those historical analyses remain documented but not independently recomputable from the release dataset.

\nocite{chollet2019measure,openai_twosettings}

\bibliographystyle{plainnat}
{\small\bibliography{references}}

\end{document}